\documentclass[runningheads]{llncs}

\usepackage{eccvabbrv}

\usepackage{graphicx}
\usepackage{booktabs}

\usepackage[accsupp]{axessibility}  % Improves PDF readability for those with disabilities.

\usepackage{hyperref}

\begin{document}

% ---------------------------------------------------------------
% TODO REVIEW: Replace with your title
\title{LSVOS Challenge 3rd Place Report: SAM2 and Cutie based VOS} 

\author{Xinyu Liu\inst{1} \and
Jing Zhang\inst{1} \and
Kexin Zhang\inst{1} \and
Xu Liu\inst{1} \and
Lingling Li\inst{1} \\
Team: Xy-unu}

\institute{\textsuperscript{1}Intelligent Perception and Image Understanding Lab, Xidian University}

\maketitle

\begin{abstract}
Video Object Segmentation (VOS) presents several challenges, including object occlusion and fragmentation, the dis-appearance and re-appearance of objects, and tracking specific objects within crowded scenes. In this work, we combine the strengths of the state-of-the-art (SOTA) models SAM2 and Cutie to address these challenges. Additionally, we explore the impact of various hyperparameters on video instance segmentation performance. Our approach achieves a J\&F score of 0.7952 in the testing phase of LSVOS challenge VOS track, ranking third overall.
\end{abstract}
\section{Introduction}
\label{sec:intro}
\begin{figure}[!htbp]
    \centering
    \includegraphics[width=\textwidth]{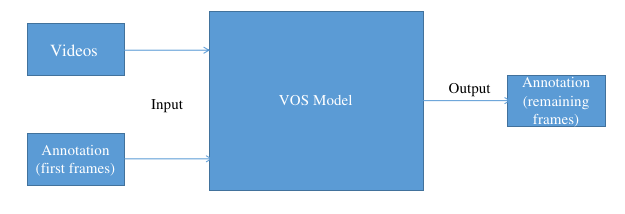}
    \caption{An overview of the VOS framework. The figure illustrates the key components of our approach, including the memory-based paradigm, pixel-level matching, and object query mechanism.}
    \label{fig:vos_framework}
\end{figure}

Video Object Segmentation (VOS) involves the identification and segmentation of target objects throughout a video sequence, starting with mask annotations in the first frame. This task is crucial in various domains, including autonomous driving, augmented reality, and interactive video editing, where the volume of video content is rapidly increasing. However, VOS faces significant challenges, such as drastic variations in object appearance, occlusions, and identity confusion caused by similar objects and background clutter. These issues are particularly challenging in long-term videos, where maintaining accurate tracking and segmentation becomes even more difficult. VOS techniques are also widely used in robotics, video editing, and data annotation, and can be integrated with Segment Anything Models (SAMs) \cite{kirillov2023segany} for universal video segmentation.Figure~\ref{fig:vos_framework} illustrates the general framework of our VOS approach. The process begins with a memory module that stores segmented frames, followed by pixel-level matching and the use of object queries to ensure accurate segmentation across all frames, even in complex scenarios.

Recent Video Object Segmentation (VOS) methods predominantly utilize a memory-based approach. These methods compute a memory representation from previously segmented frames, whether provided as input or generated by the model itself. New query frames then access this memory to retrieve features essential for segmentation. Typically, these approaches rely on pixel-level matching during memory retrieval, where each query pixel is independently matched to a combination of memory pixels, often through an attention mechanism. However, this pixel-level matching can lack high-level consistency and is vulnerable to noise, particularly in challenging scenarios with frequent occlusions and distractors. This limitation is evident in the significantly lower performance of these methods on more complex datasets like MOSE, where they can score over 20 points lower in J\&F compared to simpler datasets like DAVIS-2017. While there are methods specifically designed for VOS in long videos, they often compromise segmentation quality by compressing high-resolution features during memory insertion, leading to less accurate segmentations. In general, VOS techniques achieve segmentation by comparing test frames with previous frames, generating pixel-wise correlated features, and predicting target masks. Some approaches also employ memory modules to adapt to variations in target appearances over time and utilize object queries to differentiate between multiple objects, thereby reducing identity confusion.

This year's LSVOS Challenge features two tracks: the Referring Video Object Segmentation (RVOS) Track and the Video Object Segmentation (VOS) Track. The RVOS track has been upgraded from the Refer-Youtube-VOS dataset to the newly introduced MeViS\cite{MeViS} dataset, which presents more challenging motion-guided language expressions and complex video scenarios. Similarly, the VOS track now utilizes the MOSE\cite{MOSE} dataset, replacing the previous Youtube-VOS dataset. MOSE introduces more complexity with scenes involving disappearing and reappearing objects, small and hard-to-detect objects, and heavy occlusions, making this year's challenge significantly tougher than before. LVOS\cite{hong2023lvos,hong2024lvos}, on the other hand, focuses on long-term videos, characterized by intricate object movements and extended reappearances. As a participant in the Video Object Segmentation (VOS) track, we are required to segment specific object instances across entire video sequences based solely on the first-frame mask, further pushing the boundaries of VOS in complex environments.

\section{Method}

Our approach is inspired by recent advancements in video object segmentation, specifically the SAM 2: Segment Anything in Images and Videos by Meta \cite{ravi2024sam2} and the Cutie framework by Cheng et al \cite{cheng2023putting}.

SAM2 is a unified model designed for both image and video segmentation, where an image is treated as a single-frame video. It generates segmentation masks for the object of interest, not only in single images but also consistently across video frames. A key feature of SAM2 is its memory module, which stores information about the object and past interactions. This memory allows SAM2 to generate and refine mask predictions throughout the video, leveraging the stored context from previously observed frames.

The Cutie framework, on the other hand, operates in a semi-supervised video object segmentation (VOS) setting. It begins with a first-frame segmentation and then sequentially processes the following frames. Cutie is designed to handle challenging scenarios by combining high-level top-down queries with pixel-level bottom-up features, ensuring robust video object segmentation. Moreover, Cutie extends masked attention mechanisms to incorporate both foreground and background elements, enhancing feature richness and ensuring a clear semantic separation between the target object and distractors. Additionally, Cutie constructs a compact object memory that summarizes object features over the long term. During the querying process, this memory is retrieved as a target-specific object-level representation, which aids in maintaining segmentation accuracy across the video.

\begin{figure}[htbp]
    \centering
    \includegraphics[width=0.8\textwidth]{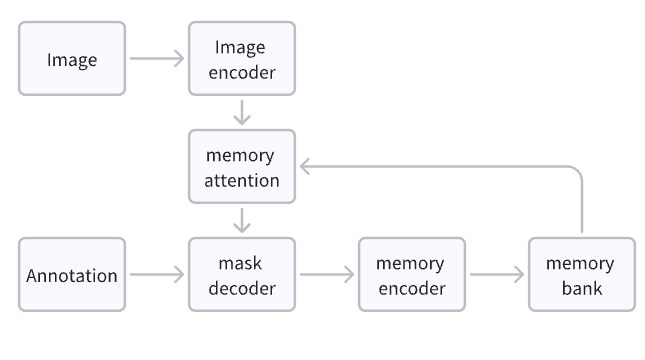}
    \caption{The SAM 2 architecture}
    \label{fig:sam2}
\end{figure}

\begin{figure}[!htbp]
    \centering
    \includegraphics[width=0.8\textwidth]{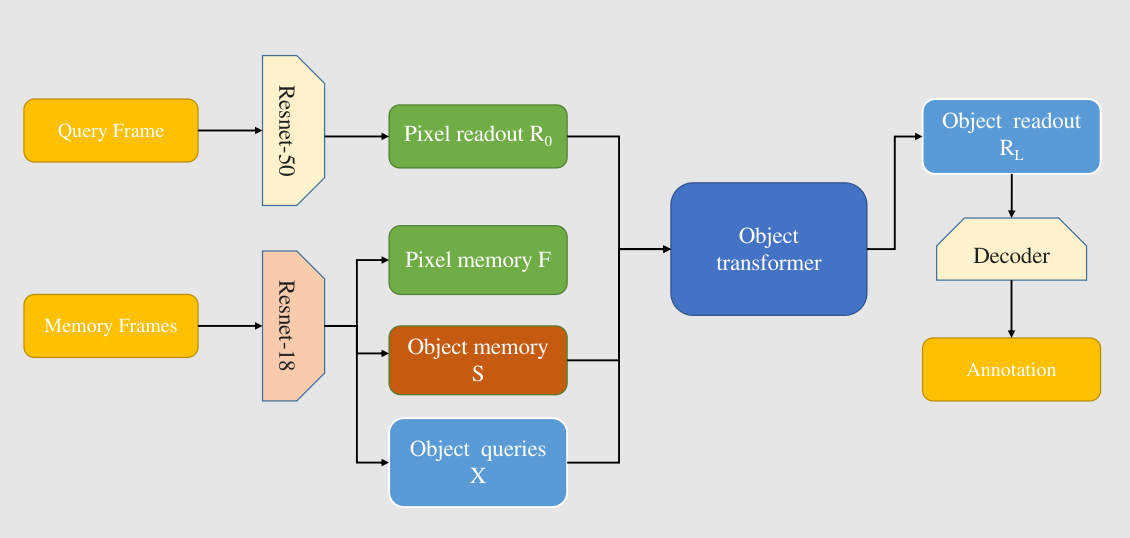}
    \caption{The Cutie architecture}
    \label{fig:cutie}
\end{figure}

As shown in Figure~\ref{fig:sam2}, the SAM2 model uses a memory-based approach for video segmentation, while Figure~\ref{fig:cutie} demonstrates how the Cutie framework incorporates object queries for improved accuracy.

\section{Experiment}
\subsection{Inference}

In our inference pipeline, we utilized the SAM2 model, specifically the sam2-hiera-large variant, which balances model size and speed effectively with a size of 224.4M and a frame rate of 30.2 FPS. To optimize the model’s performance, we compiled the image encoder by setting compile\_image\_encoder: True in the configuration. This allowed for efficient processing of high-resolution input images with a size of 1024x1024 pixels.

The configuration included several key settings designed to enhance the model’s segmentation capabilities. We utilized a num\_maskmem of 7, which refers to the number of memory tokens used in the mask memory. We also applied a scaled sigmoid function on the mask logits for the memory encoder, with parameters sigmoid\_scale\_for\_mem\_enc: 20.0 and sigmoid\_bias\_for\_mem\_enc: -10.0, to adjust the memory encoding process. Additionally, by setting use-mask-input-as-output-without-sam: true, the model directly outputs the input mask as the final mask in scenarios without SAM.

For memory management, the configuration directly\_add\_no\_mem\_embed: true ensures that new frames are directly added to the memory without additional embedding. We enabled use\_high\_res\_features\_in\_sam: true to incorporate high-resolution features into the SAM mask decoder, which improves the accuracy of mask predictions. Moreover, multimask\_output\_in\_sam: true allowed the model to output three masks upon the first interaction, enhancing the initial conditioning of frames.

The model also leverages advanced object tracking and occlusion prediction strategies. For instance, by setting use\_obj\_ptrs\_in\_encoder: true, we enabled cross-attention to object pointers from other frames during the encoding process. This, combined with pred\_obj\_scores: true and fixed\_no\_obj\_ptr: true, facilitated robust object occlusion prediction and tracking across the video sequence. Furthermore, we adopted multimask tracking settings, such as multimask\_output\_for\_tracking: true, to refine the segmentation accuracy over time.

In parallel, the Cutie framework was configured with a focus on efficient video segmentation at a standard testing resolution of 720p. In the context of the memory frame encoding, we update both the pixel memory and the object memory every $r$-th frame. The default value of $r$ is set to 3, following the same configuration used in the XMem framework~\cite{cheng2022xmem}.For subsequent memory frames, we employ a First-In-First-Out (FIFO) strategy, which ensures that the most recent information is retained while older data is gradually phased out. This approach is designed to keep the memory footprint manageable and focused on the most relevant frames.The choice of a predefined limit of $T_{\max} = 10$ for the total number of memory frames is a practical compromise. This value balances the need to avoid excessive memory usage and maintain real-time performance while still capturing sufficient temporal evolution of the scene. Maintaining a history of 15 frames is generally adequate for effectively exploiting temporal correlations in VOS tasks. This enhances segmentation accuracy by providing enough context for object tracking and appearance prediction without imposing excessive computational overhead or compromising system responsiveness. Extending this limit further could lead to diminishing returns, as the additional frames may not significantly improve performance and could increase computational load unnecessarily.
In the final testing phase, we employed Test-Time Augmentation (TTA) to enhance the model's robustness and accuracy. Specifically, we utilized flip-based augmentation, which involves horizontally flipping the input frames during inference. This simple yet effective technique helped mitigate potential overfitting and improved the model's generalization by allowing it to account for possible variations in object orientation. In the dynamic nature of video data, where frames may exhibit significant variations due to camera and object movement, flip-based TTA provided a more consistent and reliable segmentation across the video sequence.

\subsection{Evaluation Metrics}

To evaluate the performance of our model, we compute the Jaccard value (J), the F-Measure (F), and the mean of J and F.

Jaccard Value (J).
The Jaccard value, also known as Intersection over Union (IoU), measures the similarity between two sets. For a predicted segmentation mask $P$ and a ground truth segmentation mask $G$, the Jaccard value is defined as:
\begin{equation}
    J = \frac{|P \cap G|}{|P \cup G|} = \frac{\sum_{i} P_i \cdot G_i}{\sum_{i} P_i + \sum_{i} G_i - \sum_{i} P_i \cdot G_i},
\end{equation}
where $P_i$ and $G_i$ denote the value of the $i$-th pixel in the predicted and ground truth masks, respectively. The Jaccard value ranges from $0$ to $1$, with higher values indicating better performance.

F-Measure (F).
The F-Measure is a metric that combines Precision and Recall, commonly used to evaluate the performance of binary classification models. It is calculated as follows:
\begin{equation}
    F = \frac{2 \cdot \text{Precision} \cdot \text{Recall}}{\text{Precision} + \text{Recall}},
\end{equation}
where
\begin{equation}
    \text{Precision} = \frac{|P \cap G|}{|P|} = \frac{\sum_{i} P_i \cdot G_i}{\sum_{i} P_i},
\end{equation}
and
\begin{equation}
    \text{Recall} = \frac{|P \cap G|}{|G|} = \frac{\sum_{i} P_i \cdot G_i}{\sum_{i} G_i}.
\end{equation}
The F-Measure also ranges from $0$ to $1$, with higher values indicating better model performance in handling positive and negative samples.

Mean of J and F.
To comprehensively evaluate the model's performance, we compute the mean of the Jaccard value (J) and the F-Measure (F):
\begin{equation}
    \text{Mean(J, F)} = \frac{J + F}{2}.
\end{equation}

These metrics together provide a robust assessment of the segmentation model's accuracy and consistency, offering insights into its performance in predicting segmentation masks.

In the 6th Large-Scale Video Object Segmentation (LSVOS) challenge, our method (Xy-unu) demonstrated significant performance improvements in both the development and test phases. The leaderboards for these phases are presented in Tables \ref{tab:develop}, respectively. Our method achieved Jaccard values (J) and F-Measures (F) that outperformed most other participants. Specifically, in the test phase, our method attained a Jaccard value of 0.7952 and an F-Measure of 0.7516, resulting in a combined J\&F score of 0.8388.  These results highlight the effectiveness and robustness of our approach.

\begin{figure}[!htbp]
\centering
\begin{tabular}{c c c c c}
    \includegraphics[width=0.18\textwidth]{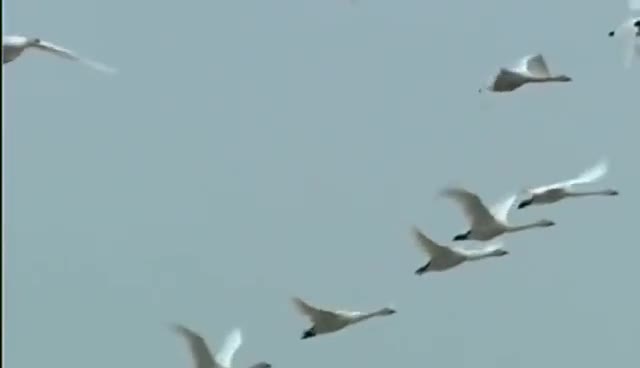} &
    \includegraphics[width=0.18\textwidth]{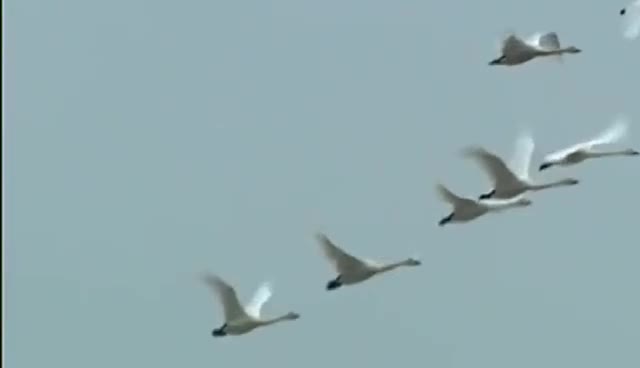} &
    \includegraphics[width=0.18\textwidth]{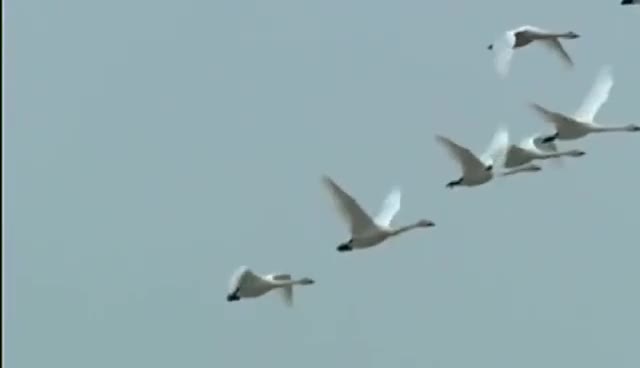} &
    \includegraphics[width=0.18\textwidth]{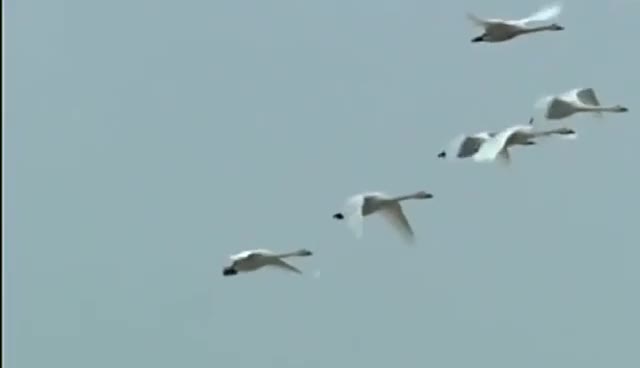} &
    \includegraphics[width=0.18\textwidth]{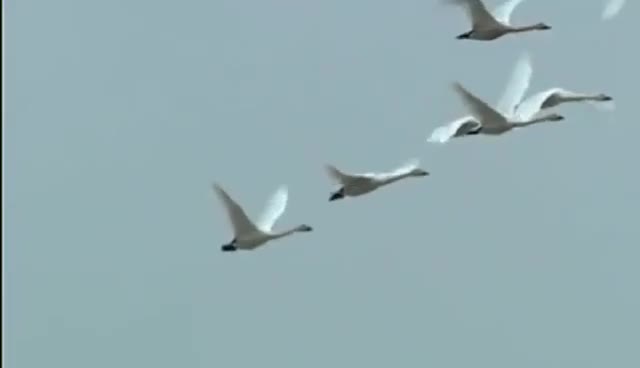} \\
    \includegraphics[width=0.18\textwidth]{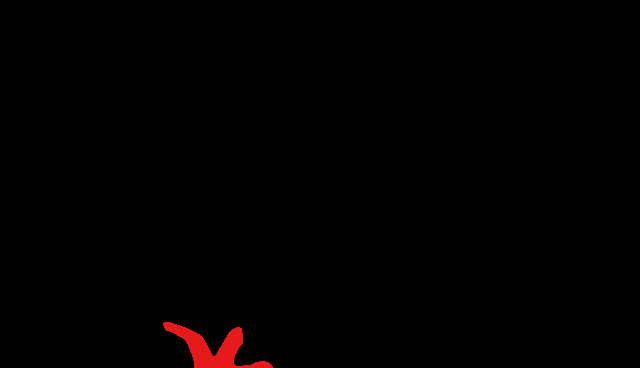} &
    \includegraphics[width=0.18\textwidth]{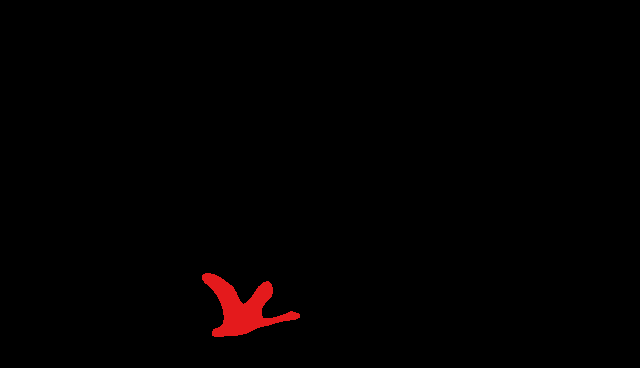} &
    \includegraphics[width=0.18\textwidth]{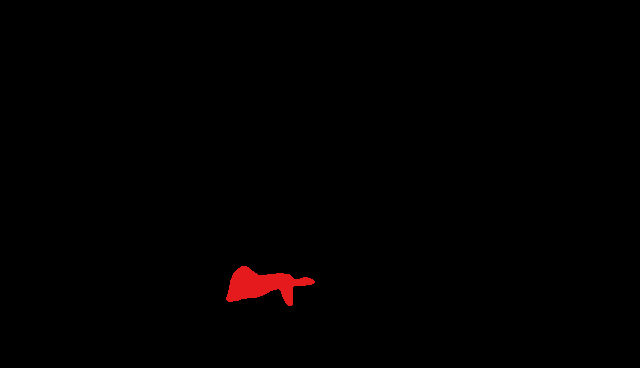} &
    \includegraphics[width=0.18\textwidth]{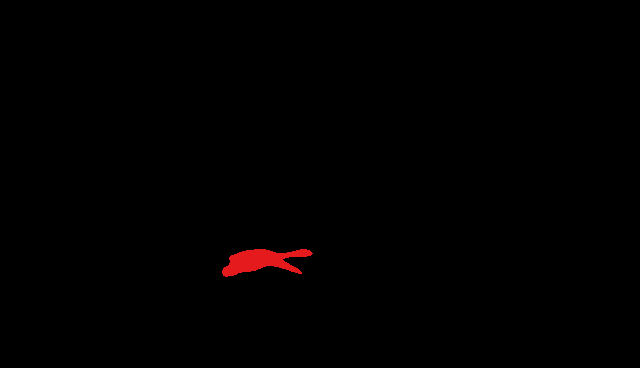} &
    \includegraphics[width=0.18\textwidth]{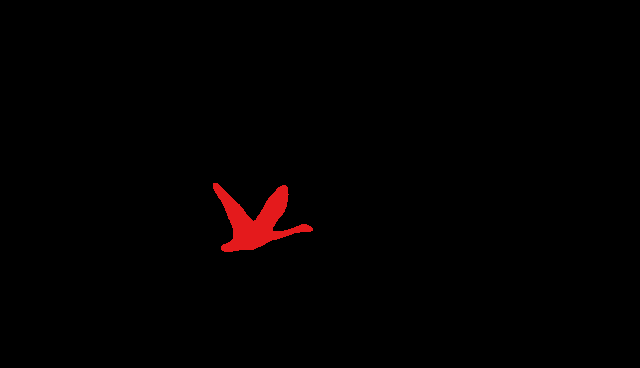} \\
    \texttt{10d24fa6 \#0 GT} & \texttt{\#3} & \texttt{\#5} & \texttt{\#7} & \texttt{\#9} \\
\end{tabular}
\caption{Performance on sequences with small targets.}
\label{fig:small_targets}
\end{figure}

Moreover, we present some of our quantitative results in Fig\ref{fig:small_targets}. The results clearly demonstrate that our proposed method is capable of accurately segmenting small targets and differentiating between similar objects in challenging scenarios. These scenarios include significant variations in object appearance and instances where multiple similar objects or small objects cause confusion.

\begin{table}[!htbp]
\setlength{\tabcolsep}{3mm}
\centering
\caption{Leaderboard during the development phase.}
\label{tab:develop}
\begin{tabular}{l|ccc}
    \toprule
    User & J & F & J\&F \\
    \midrule
    yahooo & 0.8090 (1) & 0.7616 (2) & 0.8563 (1) \\
    yuanjie & 0.8084 (2) & 0.7642 (1) & 0.8526 (3) \\
    \textbf{Xy-unu} & 0.7952 (4) & 0.7516 (4) & 0.8388 (4)  \\
    Sch89.89 & 0.7635 (7) & 0.7194 (7) & 0.8076 (7) \\
    Phan & 0.7579 (9) & 0.7125 (10) & 0.8033 (9) \\
    \bottomrule
\end{tabular}
\end{table}

\section{Conclusion}
In this work, we propose a video object segmentation (VOS) inference solution that integrates the SAM2 and Cutie frameworks. Our approach leverages the strengths of both models to handle video data effectively.Our solution demonstrated its effectiveness in the LVOS challenge, achieving a notable J\&F score of 0.7952, which secured us the third place. This result underscores the robustness and accuracy of our method in handling complex video segmentation tasks.

\clearpage  % TODO REVIEW/FINAL: This \clearpage needs to be removed from both review and camera-ready versions.

% ---- Bibliography ----
%
% BibTeX users should specify bibliography style 'splncs04'.
% References will then be sorted and formatted in the correct style.
%
\bibliographystyle{splncs04}
\bibliography{main}

\begin{thebibliography}{1}
\providecommand{\url}[1]{\texttt{#1}}
\providecommand{\urlprefix}{URL }
\providecommand{\doi}[1]{https://doi.org/#1}

\bibitem{cheng2023putting}
Cheng, H.K., Oh, S.W., Price, B., Lee, J.Y., Schwing, A.: Putting the object back into video object segmentation. In: arXiv (2023)

\bibitem{cheng2022xmem}
Cheng, H.K., Schwing, A.G.: {XMem}: Long-term video object segmentation with an atkinson-shiffrin memory model. In: ECCV (2022)

\bibitem{MeViS}
Ding, H., Liu, C., He, S., Jiang, X., Loy, C.C.: {MeViS}: A large-scale benchmark for video segmentation with motion expressions. In: ICCV (2023)

\bibitem{MOSE}
Ding, H., Liu, C., He, S., Jiang, X., Torr, P.H., Bai, S.: {MOSE}: A new dataset for video object segmentation in complex scenes. In: ICCV (2023)

\bibitem{hong2023lvos}
Hong, L., Chen, W., Liu, Z., Zhang, W., Guo, P., Chen, Z., Zhang, W.: Lvos: A benchmark for long-term video object segmentation. In: Proceedings of the IEEE/CVF International Conference on Computer Vision. pp. 13480--13492 (2023)

\bibitem{hong2024lvos}
Hong, L., Liu, Z., Chen, W., Tan, C., Feng, Y., Zhou, X., Guo, P., Li, J., Chen, Z., Gao, S., et~al.: Lvos: A benchmark for large-scale long-term video object segmentation. arXiv preprint arXiv:2404.19326  (2024)

\bibitem{kirillov2023segany}
Kirillov, A., Mintun, E., Ravi, N., Mao, H., Rolland, C., Gustafson, L., Xiao, T., Whitehead, S., Berg, A.C., Lo, W.Y., Doll{\'a}r, P., Girshick, R.: Segment anything. arXiv:2304.02643  (2023)

\bibitem{ravi2024sam2}
Ravi, N., Gabeur, V., Hu, Y.T., Hu, R., Ryali, C., Ma, T., Khedr, H., R{\"a}dle, R., Rolland, C., Gustafson, L., Mintun, E., Pan, J., Alwala, K.V., Carion, N., Wu, C.Y., Girshick, R., Doll{\'a}r, P., Feichtenhofer, C.: Sam 2: Segment anything in images and videos. arXiv preprint arXiv:2408.00714  (2024), \url{https://arxiv.org/abs/2408.00714}

\end{thebibliography}
\end{document}